\documentclass{article} 
\usepackage{iclr2024_conference,times}

\usepackage{amsmath,amsfonts,bm}

\def\eqref#1{equation~\ref{#1}}

\def\1{\bm{1}}

\DeclareMathAlphabet{\mathsfit}{\encodingdefault}{\sfdefault}{m}{sl}
\SetMathAlphabet{\mathsfit}{bold}{\encodingdefault}{\sfdefault}{bx}{n}

\usepackage{graphicx}
\usepackage{hyperref}
\usepackage{url}
\usepackage{subcaption}
\usepackage{amsmath}
\usepackage{booktabs}

\usepackage{multirow}

\newcommand*{\affaddr}[1]{#1} 
\newcommand*{\affmark}[1][*]{\textsuperscript{#1}}
\newcommand*{\email}[1]{\texttt{#1}}
\title{GNeRP: Gaussian-guided Neural Reconstruction of Reflective Objects with Noisy Polarization Priors}

\iclrfinalcopy
\author{Yang LI\affmark[1,2,4] \hfill Ruizheng WU\affmark[4] \hfill Jiyong LI\affmark[3] \hfill Yingcong CHEN\affmark[1,2, $\dagger$] \hfill \ \ \  \\ \\
\affaddr{\affmark[1]AI Thrust, HKUST(GZ), Nansha, Guangzhou, China} \\
\affaddr{\affmark[2]Department of Computer Science \& Engineering, HKUST, Clear Water Bay, HongKong SAR, China} \\
\affaddr{\affmark[3]Department of Computer Science, Sun Yat-sen University, Panyu, Guangzhou, China} \\
\affaddr{\affmark[4]R \& D Center, SmartMore, Qianhai, Shenzhen, China} \\
\affaddr{\affmark[$\dagger$] Corresponding Author}\\
\small{\email{yli803@connect.ust.hk, rzwu@cse.cuhk.hk, lijy373@mail2.sysu.edu.cn}},\\
\small{\email{yingcongchen@ust.hk}}}

\begin{document}

\maketitle

\begin{abstract}
Learning surfaces from neural radiance field (NeRF) became a rising topic in Multi-View Stereo (MVS).  Recent Signed Distance Function (SDF)--based methods demonstrated their ability to reconstruct accurate 3D shapes of Lambertian scenes. However, their results on reflective scenes are unsatisfactory due to the entanglement of specular radiance and complicated geometry. To address the challenges, we propose a Gaussian-based representation of normals in SDF fields. Supervised by polarization priors, this representation guides the learning of geometry behind the specular reflection and captures more details than existing methods.  Moreover, we propose a reweighting strategy in the optimization process to alleviate the noise issue of polarization priors. To validate the effectiveness of our design, we capture polarimetric information, and ground truth meshes in additional reflective scenes with various geometry. We also evaluated our framework on the PANDORA dataset. Comparisons prove our method outperforms existing neural 3D reconstruction methods in reflective scenes by a large margin. Supplemental materials can be found in \hyperlink{https://yukiumi13.github.io/gnerp_page/}{this page}.
\end{abstract}

\section{Introduction}
\label{sec:intro}

Reconstructing 3D shapes from 2D images \citep{mvstutor} is a fundamental problem in computer vision and graphics, with downstream applications such as 3D printing \citep{3dprint}, autonomous driving \citep{mvsdrive}, and Computer Aided Design \citep{mvsdesign}. Although diffuse objects are precisely reconstructed, reflective and textured-less scenes remain challenging. Traditional Multi-View Stereo (MVS) methods \citep{shapefromimg} rely on stereo matching across views, which is hindered in the presence of specular surfaces and texture absence. Recent methods utilizing implicit neural representation learning for 3D reconstruction have shown promising accuracy \citep{occunet, volsdf}, yet they overlook the specular reflection between light rays and surfaces, failing to adequately handle glossy objects with high-frequency specular reflection.

\begin{figure}[h]
    \centering
    \subcaptionbox{Neural 3D Reconstruction}{\includegraphics[height=0.2\textwidth]{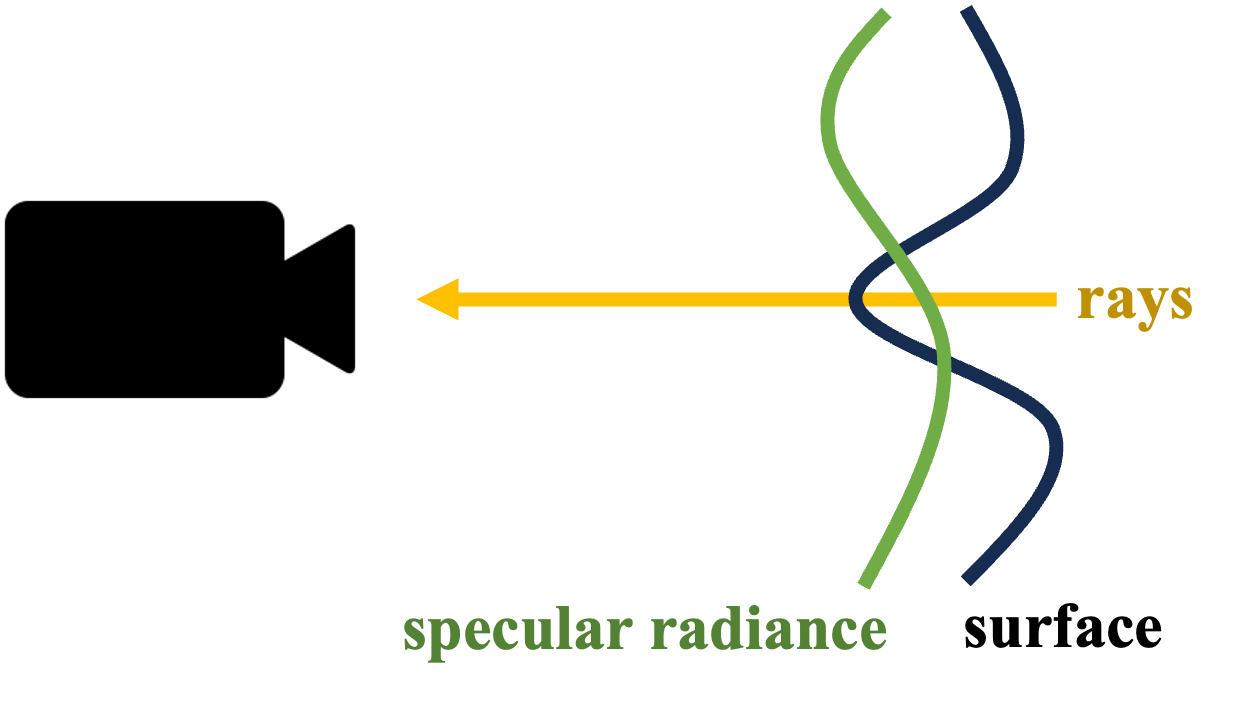}}
    \subcaptionbox{GNeRP}{\includegraphics[height=0.2\textwidth]{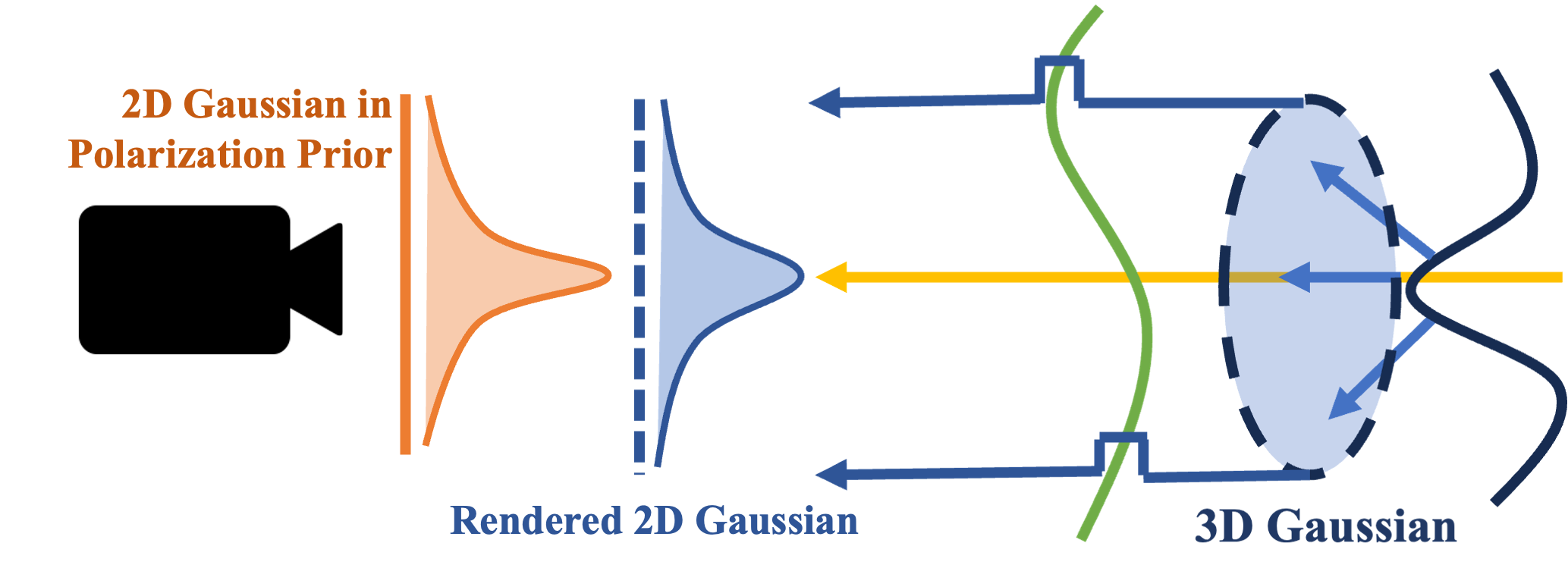}}
    \caption{Visualization of Gaussians of normals in Neural Reconstruction pipelines. 2D Gaussians can be rendered from 3D Gaussians of learned normals.}
    \label{fig:3dgaussian}
\end{figure}

Existing methods \citep{physg, nero, pandora} attempt to separate specular reflection components from radiance to improve the reconstruction process. These methods model the interaction of light rays and surfaces by Bidirectional Reflectance Distribution Functions (BRDFs) and estimate them by neural networks. However, the inverse problem posed by BRDFs formulation is highly ill-posed \citep{separate}, and low-frequency bias \citep{lowfrequency} of neural BRDFs making the learned geometry over-smoothed \citep{nero}. Therefore, high-frequency geometry with specular reflection shown in Fig. \ref{fig:3dgaussian} (a) is intractable for them. Besides, a few methods employ polarization priors to facilitate the learning of specular reflection because they reveal information about surface normals. However, polarization information is always concentrated in specular-dominant regions and noisy in diffuse regions \citep{dopsepcular}, making the reconstruction process in diffuse-dominant regions distorted.

Faced with the bias of neural BRDF and noise issues of polarization priors, we present a novel perspective for reconstructing the detailed geometry of reflective objects. Our key idea is to extend the geometry representation from scalar SDFs to Gaussian fields of normals supervised by polarization priors. Given a surface point, the normals within its neighborhood are approximated by a 3D Gaussian. And it's a more informative representation of geometry. The mean shows the overall (low-frequency) orientation of the surface, while the covariance captures high-frequency details. Coincidentally, the representation can be splatted into the image plane as 2D Gaussians, as illustrated in Fig. \ref{fig:3dgaussian} (b). The splatting skips the disentangled specular radiance. Learning of the 2D Gaussians can be directly supervised by the polarization information about surface normals. Hence, it circumvents the separation of complex geometry and specular reflection and manages to learn detailed geometry.

Furthermore, to tackle the noise issues of polarization priors, we introduce a Degree of Polarization (DoP) based reweighting strategy. This strategy adaptively balances the supervision of radiance and polarization priors, enhancing the reconstructing accuracy in diffuse-dominant regions.

In summary, our contributions are as follows:
\begin{itemize}
\item We propose a novel polarization-based Gaussian representation of detailed geometry to guide the learning of geometry behind specular reflection.
\item We propose a DoP reweighing strategy to alleviate noise and imbalance distribution problems of polarization priors.
\item We collect a new challenging multi-view dataset consisting of both radiance and polarimetric images with more diverse and challenging scenes.
\end{itemize}

\section{Related Work}
\label{sec:related}
\subsection{Multi-view 3D Reconstruction}\label{sec:mvs}
Traditional Multi-view Stereo focuses on the extraction of cross-view features to generate 3D points. \citep{schonberger2016pixelwise,galliani2015massively} try to estimate the depth map of the observed scene with multi-view consistency and fuse the depth maps into dense point clouds. These methods suffer from accumulating errors due to complex pipelines, and features are hard to be extracted from reflective objects. \citep{occunet} explicitly models the objects' occupancy in a voxel grid to guarantee a complete object model is created. However, the resolution of the voxel limits the accuracy of the reconstructed surface.
Recently, the success of NeRF \citep{nerf}, which uses a simple MLP to encode the color and density information for a scene, inspired researchers to resort to implicit representation for multi-view 3D reconstruction. The representative works are Unisurf \citep{unisurf}, NeuS \citep{neus}, and VolSDF \citep{volsdf}, which exploit an MLP to model a Signed Distance Function (SDF) for a target scene. These methods optimize the implicit representation, i.e., SDF, by minimizing the MSE loss between the rendered pixel's radiance value and the corresponding pixel's radiance value in GT images. Such a paradigm works well with Lambertian surfaces. However, only view-conditioned radiance fields fail in reflective scenes. 
\subsection{BRDF for Reflective Objects Reconstruction}\label{sec:brdf}
In the regions with complex geometry, BRDFs always exhibit high-frequency variations due to the normals terms, while the low-frequency implicit bias of neural networks \citep{lowfrequency} disables neural BRDFs from predicting these abrupt changes. It always results in over-smoothed geometry. For example, NeRO \citep{nero} adopts Micro-facet BRDF \citep{microfacet} parameterized by material and normal distribution terms. Although its results of smooth mirror-like objects are excellent, the spatial continuity of neural BRDF is a barrier to the combination of complex geometry and specular reflection. In the regions with complex geometry, sole multi-view images with disentangled radiance result in severe ill-posedness of the inverse problem, as is shown in Fig. \ref{fig:3dgaussian} (a). Moreover, explicit estimation of anisotropic normals distribution has been used in rendering delicate objects, such as anisotropy shading of hairs \citep{hairrender}, to improve the perception of orientation and shapes \citep{anisotropicvolrender}. However, anisotropic normals distribution in neural SDFs for 3D reconstruction remains under-defined and non-trivial. Our method proposes 3D Gaussians, of which anisotropic 3D covariance is more informative than the scalar normals distribution term in NeRO. The latter only measures the concentration of normals at a surface point.
\subsection{Multi-view 3D Reconstruction with Polarization}\label{sec:relatepol}
Polarization prior reveals the azimuth angle of the surface normal, i.e., the angle between the normal projection onto the image plane and the positive x-axis of the image. Shape-from-polarization has been investigated by other papers \citep{recover, polvision, polnormal, pmvs, pol3d, pmvir} before the invention of neural 3D reconstruction. But most of them are focused on common scenes. For example,  PMVIR \citep{pmvir} exploits the relation of the polarization angle and the azimuth angle of normals but with only Lambert shading, and thus it cannot treat reflective objects at all. Neural 3D Reconstruction with polarization priors has also been explored. Sparse Ellipsometry \citep{ellis} develops a device to capture polarimetric information and 3D shapes concurrently. However, their reconstruction is always disturbed by the noise in diffuse-dominant regions. For example, PANDORA \citep{pandora} extends radiance in BRDF into polarimetric dimensions while the geometry of diffuse regions cannot be learned properly. 
\subsection{Gaussians in 3D Scene Representation}\label{sec:relategauss}
Gaussians are used to represent the attributes of 3D scenes. Mip-NeRF \citep{mipnerf} encodes Gaussian regions of space rather than infinitesimal points for anti-aliasing. \citep{ewa} proposes Gaussian splatting that taking volume data as 3D Gaussians and nearly projects the 3D Gaussian to the 2D one \citep{3dgaussian}. \citep{3dgaussian} implements the splatting pipeline on the NeRF for real-time rendering. In numerical geometry, \citep{diffgeo} calculates the covariance matrix from the projections of the normal vectors to highlight the edges and local geometry of surfaces. Inspired by them, we demonstrate a further fact that taking surface normals as 3D Gaussians and going through a similar splatting pipeline would exactly be transformed into 2D Gaussians. Our 2D Gaussians are coincidentally available for polarization priors. Thus, supervised by polarization priors, the learned 3D Gaussians capture more details, which represent the average orientation of normals by means and the changes within the neighborhood by covariance matrices. 

\section{Methods}
\subsection{Preliminary of Polarization}
Here, we introduce the concept of polarization and its mathematical relation to surface normals projected to the captured images. The prior contributes to the disentanglement of specular radiance and geometry.

\begin{figure}[!htb]
\centering
\captionsetup[subfloat]{labelsep=none,format=plain,labelformat=empty}
\subfloat{\includegraphics[width=0.6\textwidth]{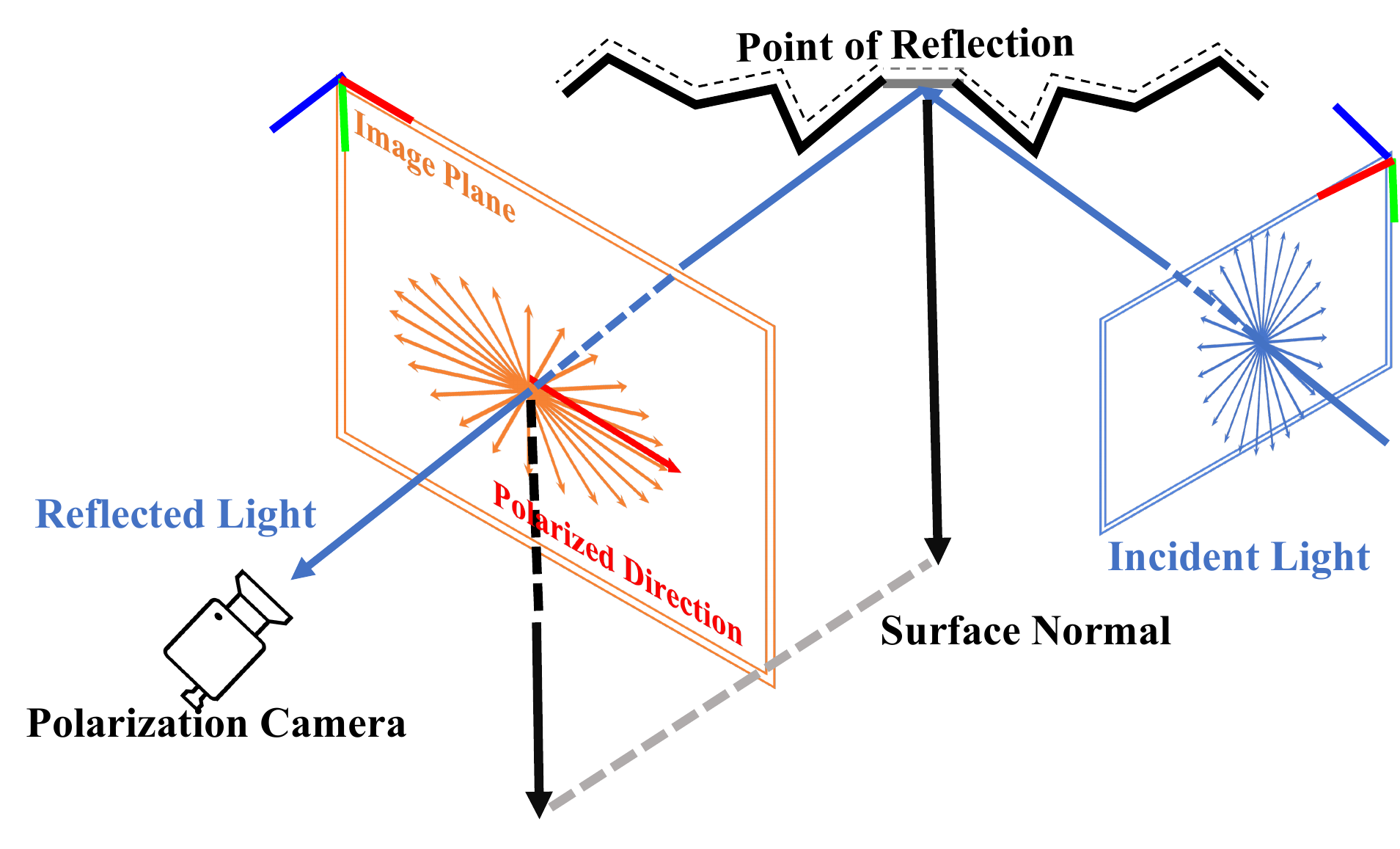}} 
\subfloat{\includegraphics[width=0.35\textwidth]{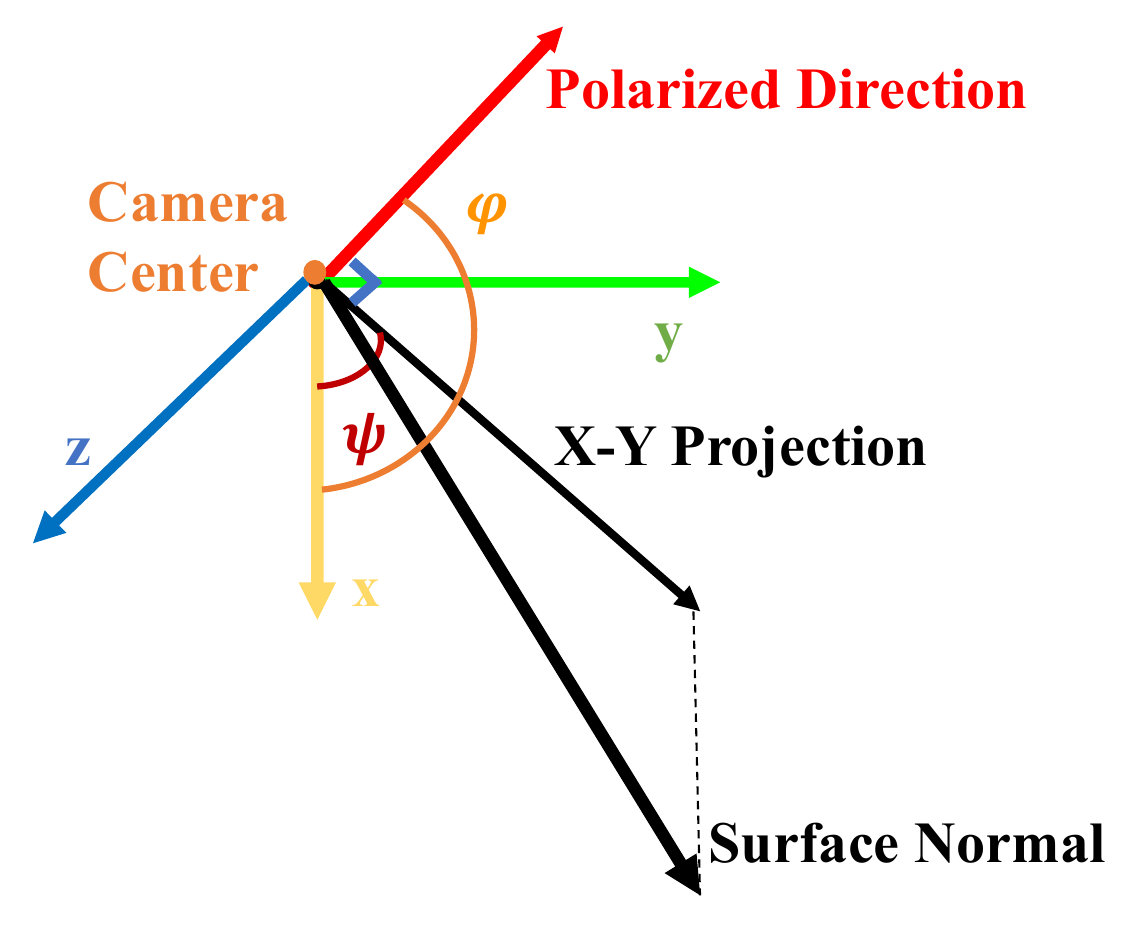}} 
\caption{Illustration of polarization shift in specular reflection. The right figure is a detailed description of the geometry relation between AoP and the surface normal. $\psi$ is the azimuth angle. $\varphi$ is the AoP, which is the angle from the positive x-axis to the polarized direction.}
\label{fig:pol_shift}
\end{figure}
Polarimetry describes the vibration status of light waves. Since light is a type of transverse wave that only oscillates in the plane perpendicular to the light path \citep{polfield}, the full polarimetric cues of rays are always represented by planar ellipses \citep{polellipse}. The magnitude of vectors inside these ellipses alludes to the amplitude of the light wave vibration along the vectors, as shown in Fig. \ref{fig:pol_shift}. Common light sources, such as sunlight and LED spotlights, emit unpolarized light, i.e., the light vibrates equally in all directions. In our captured scenes, objects are mostly illuminated directly by light sources, so we assume that the incident light is unpolarized. During reflection, the vibration in each direction is absorbed unequally, and unpolarized incident light turns into partially polarized reflected light captured by polarization cameras. The Angle of Polarization (AoP) and Degree of Polarization (DoP) are two cues of the polarization ellipse functionally related to projected surface normals at the points of reflection, which can be formulated as:
\begin{equation}\label{eq:pol}
\boldsymbol{\varphi}(i,j) = \dfrac{1}{2} \arctan{\dfrac{\mathbf{s}_2(i,j)}{\mathbf{s}_1(i,j)}},\ 
\boldsymbol{\rho}(i,j) = \dfrac{\sqrt{\mathbf{s}^2_1(i,j) + \mathbf{s}^2_2(i,j)}}{\mathbf{s}_0(i,j)}
,\ \{ \boldsymbol{\varphi},\boldsymbol{\rho}\} \in \mathbb{R}^{H \times W}, 
\end{equation} 
where $\boldsymbol{\varphi}, \boldsymbol{\rho}$ are AoP and DoP, $(i,j)$ is the pixel index, and $\mathbf{s}=[\mathbf{s}_0,\mathbf{s}_1,\mathbf{s}_2,\mathbf{s}_3]$ is Stokes vector directly calculated from polarization capture. Generally, in specularity-dominant regions, the relation between projected normals and AoP is fixed as Fig \ref{fig:pol_shift}(b) and the equation $\mathbf{\psi} + \dfrac{\pi}{2} \equiv \varphi \mod \pi$. Moreover, DoP is significantly higher in these regions. Details of polarization analysis are shown in the Appendix.

\subsection{Gaussian Guided Polarimetric Neural 3D Reconstruction Pipeline}
\begin{figure}[h]
\begin{center}
\includegraphics[width=\linewidth]{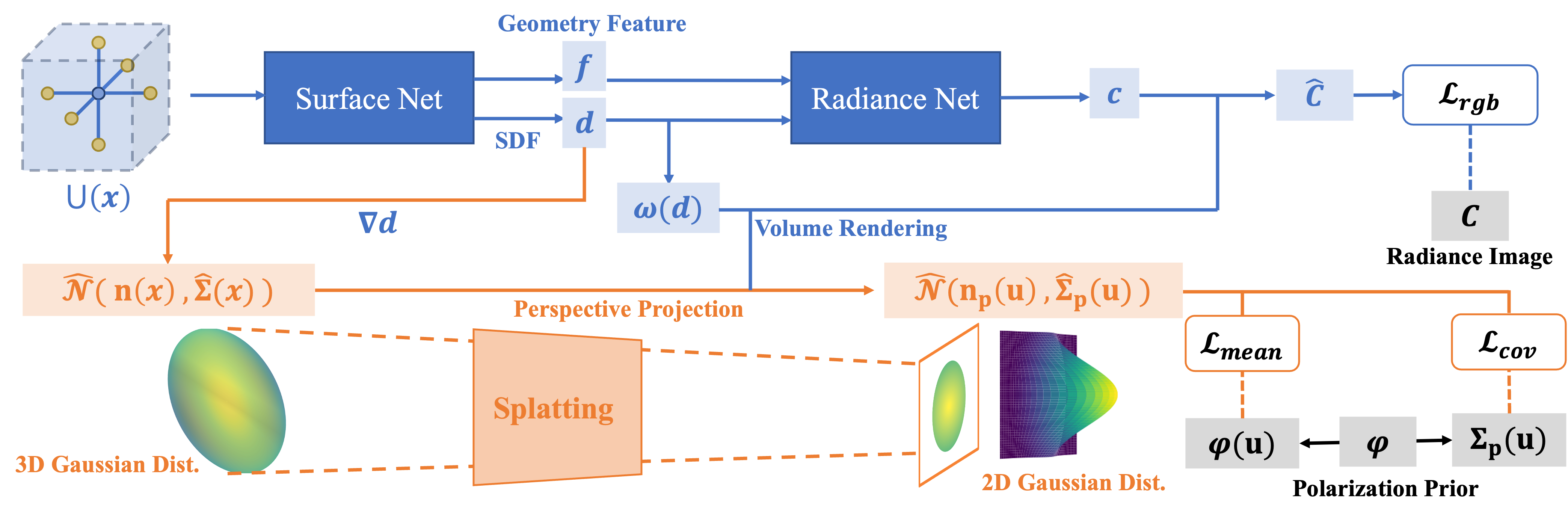}
\end{center}
   \caption{Illustration of our method.}
\label{fig:net}
\end{figure}

Polarimetric neural 3D reconstruction refers to reconstructing surfaces by neural implicit surface learning, given $N$ calibrated multi-view images $\mathcal{X}=\{ \mathbf{C}_i\}_{i=1}^N$ with pixel-aligned polarization priors $\mathcal{Y} = \{ \boldsymbol{\varphi}_i, \boldsymbol{\rho}_i \}_{i=1}^N$.  First, we introduce a general pipeline of learning surface by volume rendering, taking NeuS \citep{neus} as an example. Sec. \ref{sec:gaussian} introduces the 3D Gaussian of surface normals and its transforms to 2D Gaussian in the image plane. It represents the geometry of surface points more precisely and thus can separate detailed geometry from high-frequency specular radiance. \ref{sec:opt} presents our full optimization containing radiance loss and Gaussian loss, which measures the gap between these 2D Gaussians and polarization priors. We propose a DoP reweighing strategy to alleviate the aforementioned noise and imbalanced distribution of polarization priors. It balances the influence of radiance and polarimetric cues adaptively. Finally, Sec. An overview of the entire framework is shown in Fig. \ref{fig:net}.

\subsubsection{Learning Surface by Volume Rendering}\label{sec:volrender}
NeRF \citep{nerf} proposed a novel render pipeline with a combination of spatial neural radiance fields and volume rendering \citep{volrender} to synthesize high-quality novel view images. Unlike traditional explicit meshes, the representation of 3D scenes in NeRF is decomposed into spatial-dependent density fields and radiance fields depending on both spatial position and viewing direction. Then, the color of an arbitrary pixel with a ray $\mathbf{r} = \mathbf{o} + t \mathbf{d}$ passed through can be rendered by volume composition along the ray:
\begin{equation}\label{eq:volrender}
    \hat{\mathbf{C}}(\mathbf{r}) = \sum_{k=1}^K T_i\alpha_i \mathbf{c}_i(\mathbf{r}_i,  \mathbf{d}),\  T_i = \exp{(- \sum_{j=1}^{i-1} \alpha_j \delta_j )},\  \alpha_j = 1 - \exp{(- \sigma_j(\mathbf{r}_j)\delta_j)},
\end{equation}
, where $K$ points $\{ \mathbf{x} \vert \mathbf{o} + t_i \mathbf{d}\}_{i=1}^K$ on the ray are sampled. $\sigma_i$ and $\mathbf{c}_i$ are approximated volume density and radiance predicted by neural networks with position $\mathbf{x}$ and viewing direction $\mathbf{d}$ as inputs. $\delta_i$ is the length of sampled interval $[t_{i-1}, t_i]$. $\alpha_i$ and $T_i$ denote the transmittance and alpha value of points, and by them the final color is alpha composited \citep{alphacomposition}. The neural network is optimized by the mean square error between the ground truth color $\mathbf{C}(\mathbf{r})$  in the image and the rendered color $\hat{\mathbf{C}}(\mathbf{r})$.

Despite realistic novel view images, the geometry of scenes extracted from learned density fields is inaccurate with floating artifacts since the shape is not defined in the density field. NeuS \citep{neus} defines surfaces as the zero-level set of Signed Distance Field (SDF), and density is derived from SDF:
\begin{equation}
    [d(\mathbf{x}_i), \mathbf{f}(\mathbf{x}_i)] = f(\mathbf{x}_i),\ \alpha_i=\max \left(\frac{\Phi_s\left(d\left(\mathbf{x}_i\right)\right)-\Phi_s\left(d\left(\mathbf{x}_{i+1}\right)\right)}{\Phi_s\left(d\left(\mathbf{x}_i\right)\right)}, 0\right),\ \mathbf{c}_i = c(\mathbf{x}_i,\mathbf{n}_i, \mathbf{d}, \mathbf{f}_i),
\end{equation}
where $c$ and $f$ are the geometry network and radiance network, $d(\mathbf{x}_i$) is the signed distance to the surface and $\mathbf{f}_i = \mathbf{f}(\mathbf{x}_i)$ is the geometry feature. $\alpha_i$ is defined by SDF with Laplace distribution $\Phi_s\left( x \right) = ({1 + e^{-sx}})^{-1}$, where the variance $s$ is a trainable parameter. The volume rendering process is analogous to NeRF, while the radiance network takes normals $\mathbf{n}_i = \nabla_{\mathbf{x}} d(\mathbf{x}_i)$ and geometry feature $\mathbf{f}_i$ as additional inputs.
\subsubsection{Gaussian Splatting of Normals}\label{sec:gaussian}
Neural SDF-based 3D reconstruction excels at smooth Lambertian objects. With neural BRDF defining the specular reflection between rays and surfaces, smooth surfaces of reflective objects can also be properly learned. However, the low-frequency implicit bias of neural networks \citep{lowfrequency} is a barrier for both of them to recover delicate geometry behind specular reflection, such as abrupt normal changes in NeRO \citep{nero}. Thus, we propose a 3D Gaussian estimation of distributions of normals as an additional representation of geometry details. We show how it can be splatted to the image plane, making it available for 2D polarization supervision.

Instead of separate vectors assigned to each point, the normal within the neighborhood of an arbitrary position $\mathbf{x_i}$ is assumed as a Gaussian:
\begin{equation}
    \mathcal{G}(\mathbf{x} \vert \mathbf{x}_i) = \mathcal{N}( \mathbf{n}(\mathbf{x}_i), \mathbf{\Sigma}(\mathbf{x}_i))= \frac{1}{(2 \pi)^{\frac{3}{2}}|\mathbf{\Sigma}(\mathbf{x}_i)|^{\frac{1}{2}}} \exp{\left(-\frac{1}{2}(\mathbf{x}-\boldsymbol{\mathbf{n}(\mathbf{x}_i)})^{\mathrm{T}} \mathbf{\Sigma({x}_i)}^{-1}(\mathbf{x}-\boldsymbol{\mathbf{n}(\mathbf{x}_i)})\right)},z
\end{equation}
where $\mathbf{n} \in \mathbb{R}^3$ is the normal, and $\mathbf{\Sigma} \in \mathbb{R}^{3 \times 3}$ is the covariance of the Gaussian. Given a ray with discretization $\{\mathbf{x}_i \vert \mathbf{x} + t_i \mathbf{d} \}_{k=1}^K$, additional $M$ positions within the neighborhood are super-sampled to estimate the covariance. In this paper, $M$ is 6 containing $\mathbf{x}_{i-1}$, $\mathbf{x}_{i+1}$ and four positions around the ray. Hence, the unbiased estimation of Gaussian can be formulated as:
\begin{equation}\label{eq:estimate_3d_gaussian}
    \hat{\mathcal{G}}(\mathbf{x} \vert \mathbf{x}_i) = \mathcal{N}( \mathbf{n}(\mathbf{x}_i), \hat{\mathbf{\Sigma}}(\mathbf{x}_i)) = \mathcal{N}\left( \mathbf{n}(\mathbf{x}_i), \dfrac{1}{M-1} \sum_{j=1}^{M} \left(\mathbf{n}(\mathbf{x}_{i}^{j}) - \mathbf{n}(\mathbf{x}_{i})\right)\left(\mathbf{n}(\mathbf{x}_{i}^{j}) - \mathbf{n}(\mathbf{x}_{i})\right)^\mathrm{T}\right),
\end{equation}
where $\mathbf{n}(\mathbf{x}_i^j) = \nabla_{\mathbf{x}}d(\mathbf{x}_i^j), \mathbf{n}(\mathbf{x}_i^j) \in \mathbb{R}^{3}$. However, those 3D Gaussians are not accessible in captured 2D images, and volume rendering in Sec. \ref{sec:volrender} only takes 3D scalar fields into account, making the projection of 3D Gaussians non-trivial. Alternatively, \citep{ewa} presents a splatting approach treating colors in 3D space as Gaussian kernels and visualizing them on the image plane. We apply analogous transforms and further prove our normal-based 3D Gaussians are exactly splatted to 2D Gaussians. Given a viewpoint, the transform can be formulated as:
\begin{equation}\label{eq:proj}
    \hat{\mathcal{G}}(\mathbf{x} \vert \mathbf{x}_i)_{\mathbf{p}} = \mathcal{N}( \mathbf{J}\mathbf{W}\mathbf{n}(\mathbf{x}_i), \mathbf{J}\mathbf{W}\hat{\mathbf{\Sigma}}(\mathbf{x}_i))\mathbf{W}^{\mathrm{T}}\mathbf{J}^{\mathrm{T}}) = \mathcal{N}\left(\left[\begin{array}{c}
\mathbf{n}_{\mathbf{p}}(\mathbf{x}_i) \\
 0
\end{array}\right],\left[\begin{array}{l l}
\hat{\mathbf{\Sigma}}_{\mathbf{p}}(\mathbf{x}_i) &  \\
  & 0
\end{array}\right]\right),
\end{equation}
where $\mathbf{W} \in \mathbb{R}^{3 \times 3}$, $\mathbf{J} \in \mathbb{R}^{3 \times 3}$ are viewing transform matrix and normal projection matrix \citep{ppa}, respectively. Derivation of them is shown in the Appendix. It shows that only the first two rows of the transformed mean vector and the upper $2 \times 2$ square block of the transformed covariance matrix remain non-zero, splatting 3D Gaussians to 2D Gaussians in the image plane. For simplification, 2D Gaussians are also denoted by $\hat{\mathcal{G}}(\mathbf{x} \vert \mathbf{x}_i)_{\mathbf{p}} = \mathcal{N}(\mathbf{n}_{\mathbf{p}}(\mathbf{x}_i),\hat{\mathbf{\Sigma}}_{\mathbf{p}}(\mathbf{x}_i))$, where $\mathbf{n}_{\mathbf{p}} \in \mathbb{R}^2, \hat{\mathbf{\Sigma}}_{\mathbf{p}}(\mathbf{x}_i)) \in \mathbb{R}^{2 \times 2}$. Moreover, the SVD of the covariance matrix $\hat{\mathbf{\Sigma}}_{\mathbf{p}} = \hat{\mathbf{V}}\hat{\mathbf{\Lambda}} \hat{\mathbf{V}}^{\mathrm{T}}$ circumvents the ill-posedness of the covariance matrix and reveals its relation to anisotropic normal distribution. Intuitively, if the geometry appears smooth from the imaging perspective, then the corresponding normals of the neighborhood will be projected to similar vectors, resulting in an insignificant deviation of the eigenvalues. Otherwise, the deviation would be significant. Eigenvectors also show the local shape of the position, as shown in Fig. \ref{fig:opt} (e). Finally, all 2D Gaussians on the ray passing through the pixel $\mathbf{u}$ is composited by volume rendering:
\begin{equation}\label{eq:vol_render_gaussian}
    \hat{\mathcal{G}}(\mathbf{u}) = \mathcal{N}\left(\sum_{k=1}^K T_i\alpha_i \mathbf{n}_{\mathbf{p}}(\mathbf{x}_i), \sum_{k=1}^K T_i\alpha_i\hat{\mathbf{\Sigma}}_{\mathbf{p}}(\mathbf{x}_i)\right) = \mathcal{N}(\mathbf{n}_{\mathbf{p}}(\mathbf{u}), \hat{\mathbf{\Sigma}}_{\mathbf{p}}(\mathbf{u})),
\end{equation}
where $T_i$ and $\alpha_i$ are in Eq. \ref{eq:volrender}. $\mathbf{n}_{\mathbf{p}}(\mathbf{u}) \in \mathbb{R}^2, \hat{\mathbf{\Sigma}}_{\mathbf{p}}(\mathbf{u}) \in \mathbb{R}^{2 \times 2}$. The mean of 3D Gaussians $\mathbf{n}(\mathbf{x}_i)$, which is splatted to $\mathbf{n}_{\mathbf{p}}(\mathbf{u})$, represents the overall orientation of $\mathbf{x}_i$. And the covariance $\hat{\mathbf{\Sigma}}(\mathbf{x}_i))$ and splatted $\hat{\mathbf{\Sigma}}_{\mathbf{p}}(\mathbf{u}))$ in the image model the high-frequency details. In this way, our representation captures more details than NeuS and other methods based on the neural BRDF parameterized by isotropic normals distribution. Another main strength of those 2D Gaussians is direct supervision by polarization priors, which is introduced in Sec. \ref{sec:opt}. 
\subsubsection{ Optimization with Reweighted Polarization Priors}\label{sec:opt}
\begin{figure}[h]
\centering
\subcaptionbox*{Visualization of DoP Reweighing}{\subcaptionbox{Scene}{\includegraphics[width=0.1\textwidth]{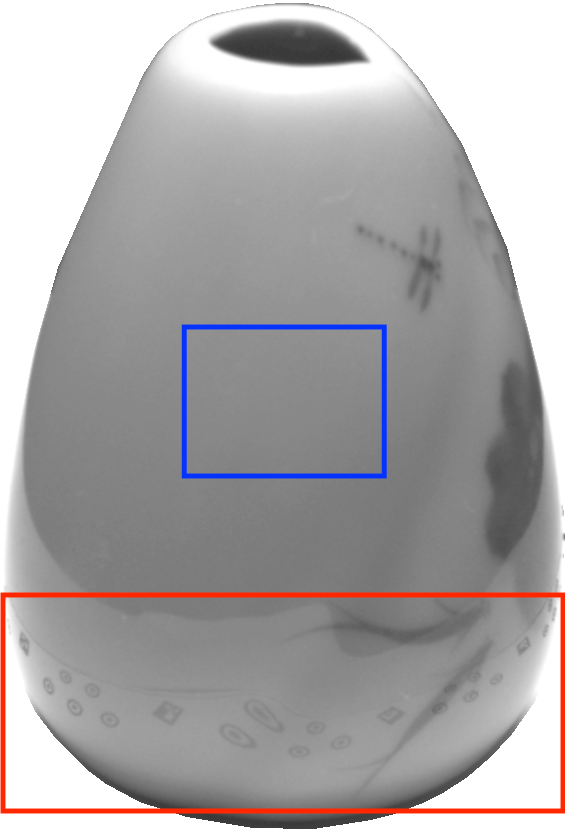}} 
\subcaptionbox{AoP}{\includegraphics[width=0.1\textwidth]{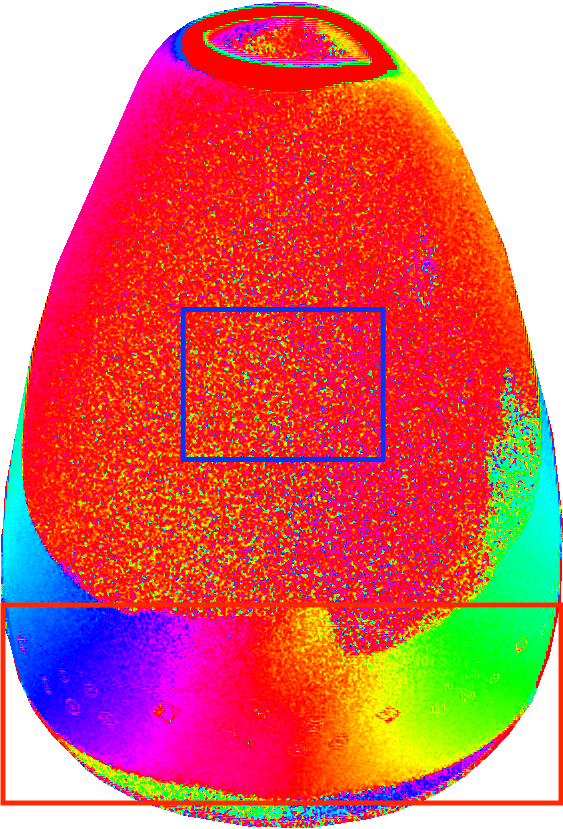}} 
\subcaptionbox{DoP}{\includegraphics[width=0.1\textwidth]{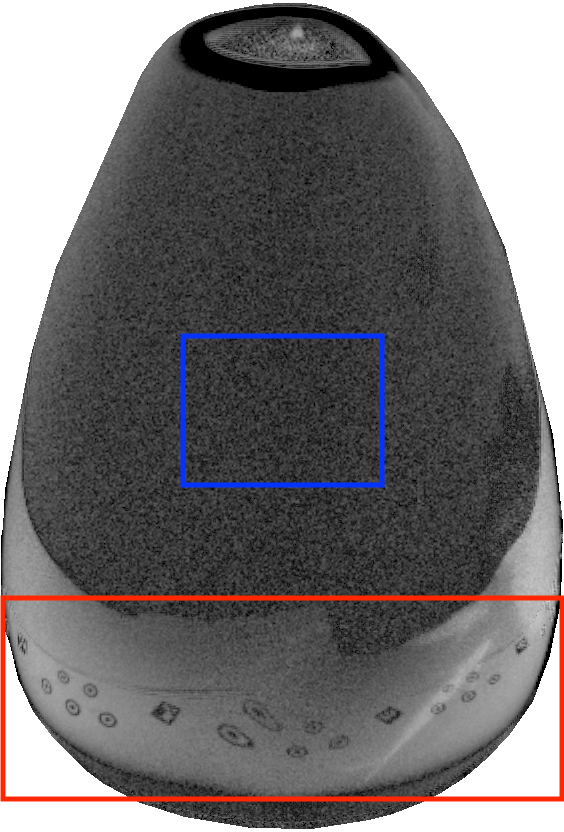}}
\subcaptionbox{R. AoP}{\includegraphics[width=0.15\textwidth]{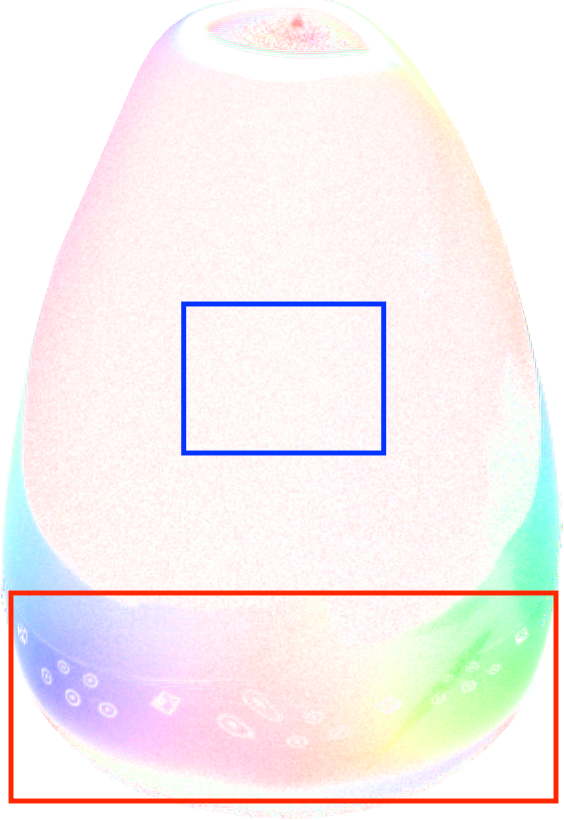}}}
\subcaptionbox{2D Gaussians in Polarization}{\includegraphics[height=0.15\textheight]{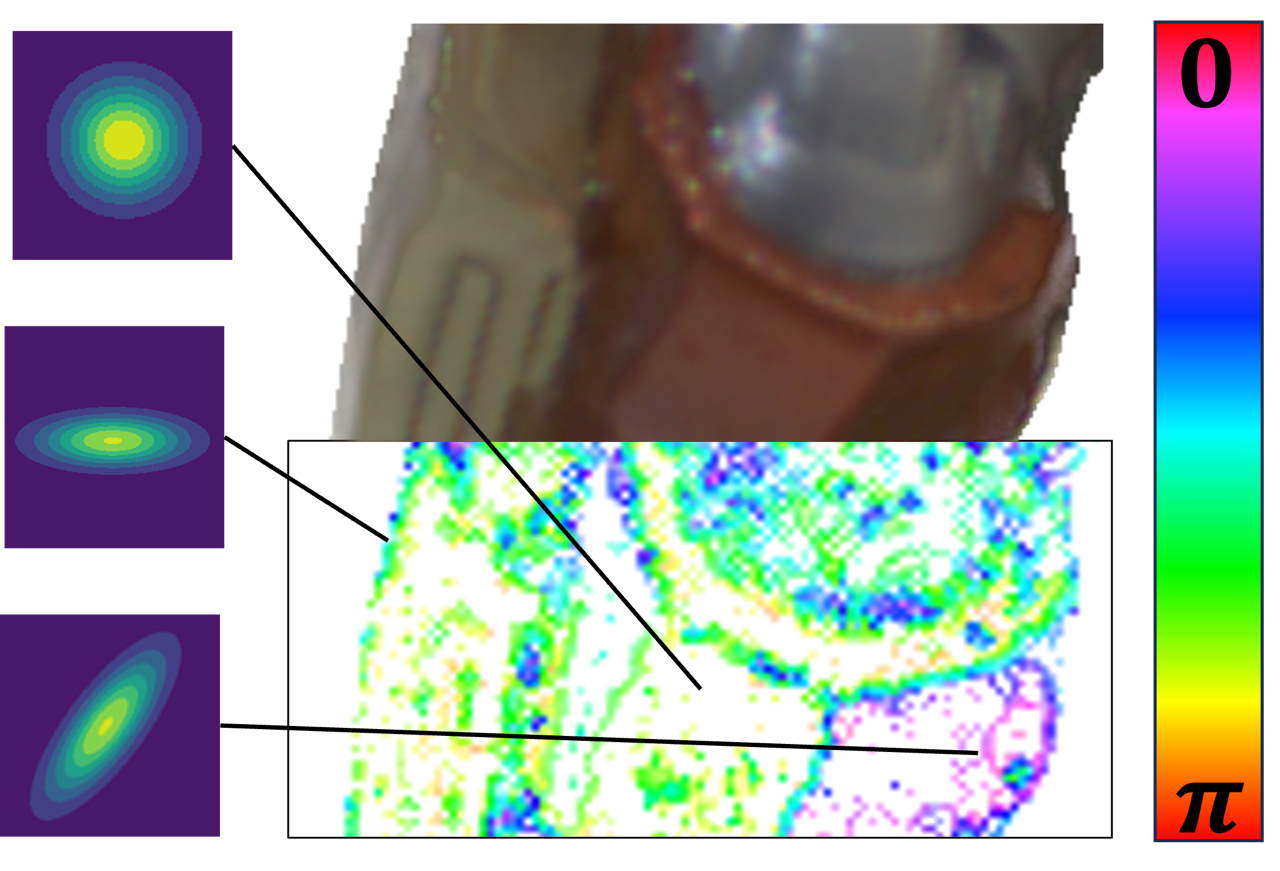}}
\caption{Visualization of Reweighted AoP Priors. Red boxes bound specular reflection dominant regions, and the blue boxes bound diffuse ones. (d) is the AoP map reweighted by DoP. Saturation in (e) indicates the degree of anisotropy, and color represents the direction of the singular vector of 2D Gaussians' covariance. A few 2D Gaussians are drawn as ellipses for intuition.}
\label{fig:opt}
\end{figure}
The 2D Gaussian in Sec. \ref{sec:gaussian} can be directly extracted from AoP priors $\{\boldsymbol{\varphi}_i\}_{i=1}^N$ in Eq. \ref{eq:pol} by:
\begin{equation}\label{eq:2dpol}
\begin{aligned}
    \widetilde{\mathcal{G}}_{\mathbf{p}}(\mathbf{u} \vert \mathbf{u}_i) =& \mathcal{N}(\widetilde{\mathbf{n}}_{\mathbf{p}}(\mathbf{u}_i), \widetilde{\mathbf{\Sigma}}_{\mathbf{p}}(\mathbf{u}_i)) \\=& \mathcal{N}\left(s\mathbf{v}(\boldsymbol{\psi}(\mathbf{u}_i)),\dfrac{1}{M'-1} \sum_{j=1}^{M'} \left(\mathbf{v}(\boldsymbol{\psi}(\mathbf{u}_i^j)) - \mathbf{v}(\boldsymbol{\psi}(\mathbf{u}_i)))\right)\left(\mathbf{v}(\boldsymbol{\psi}(\mathbf{u}_i^j)) - \mathbf{v}(\boldsymbol{\psi}(\mathbf{u}_i)))\right)^{\mathrm{T}}\right),
\end{aligned}
\end{equation}
where $\mathbf{u}_i^{(j)} = (\mathbf{x}_i^{(j)})_{\mathbf{p}}$ is the corresponding pixel index of (super-sampled) points on the ray. Therefore, $M'=4$ since $\mathbf{x}_{i-1}$ and $\mathbf{x}_{i+1}$ are on the same ray as $\mathbf{x}_i$. $\mathbf{v(\theta)}$ represents a 2D unitary vector rotated by $\theta$. $\mathbf{\Psi} \equiv \boldsymbol{\varphi} + \frac{\pi}{2} \mod \pi$ is the azimuth angle of normals, derived from the AoP in Eq. \ref{eq:pol}. And $s$ is a scale factor. Similar to Sec. \ref{sec:gaussian}, the estimated covariance matrix is decomposed into $\widetilde{\mathbf{\Sigma}} = \widetilde{\mathbf{V}}\widetilde{\boldsymbol{\Lambda}} \widetilde{\mathbf{V}}^{\mathrm{T}}$. We define the degree of anisotropy (DoA) of those 2D Gaussians as $\frac{\boldsymbol{\Lambda_0}}{\boldsymbol{\Lambda_1}}$. 2D Gaussians saturated by DoA are visualized in Fig. \ref{fig:opt} (e). In this Fig, color is concentrated to and coherent along the edges of the scene. It shows DoA is higher in the region with complicated geometry and surface changes most dramatically along singular vectors of covariance.
Before optimization, the polarization prior AoP is reweighted by DoP to alleviate the aforementioned observational noise and imbalanced distribution problem in Sec. \ref{sec:intro}. The noise of AoP is mainly generated by diffuse reflection because it's always weakly polarized \citep{dopsepcular}. The DoP in diffuse-dominant regions is significantly lower than specular-dominant ones, as shown in Fig. \ref{fig:opt} (c) and (d). Thus, the reweighted AoP defined as $\boldsymbol{\varphi \cdot \rho}$ is proposed as an alternative supervision with less noise. Meanwhile, radiance is disentangled with surroundings in specular reflection dominant areas. To adaptively balance radiance and polarization priors, our full loss function during reconstructing is defined as:
\begin{equation}\label{eq:loss}
\begin{aligned}
\mathcal{L} =& \alpha(1-\boldsymbol{\rho}) \mathcal{L}_{\mathrm{color}} + \beta \boldsymbol{\rho} (\mathcal{L}_{\mathrm{mean}} +  \mathcal{L}_{\mathrm{cov}}) + \gamma \mathcal{L}_{\mathrm{eik}} + \delta \mathcal{L}_{\mathrm{mask}}, \\
\mathcal{L}_{\mathrm{color}}=& \parallel \hat{\mathbf{C}}(\mathbf{u}) - \mathbf{C}(\mathbf{u}) \parallel_2,\  \mathcal{L}_{\mathrm{mean}}= \parallel \hat{\boldsymbol{\varphi}}(\mathbf{n}_{\mathbf{p}}(\mathbf{u})) - \boldsymbol{\varphi}(\mathbf{u}) \parallel_1,\\ \mathcal{L}_{\mathrm{cov}} =&\left(\left\Vert \frac{\hat{\boldsymbol{\Lambda}}_1}{\hat{\boldsymbol{\Lambda}}_0} - \frac{\widetilde{\boldsymbol{\Lambda}}_1}{\widetilde{\boldsymbol{\Lambda}}_0} \right\Vert_1 + \beta' <\hat{\mathbf{V}}, \widetilde{\mathbf{V}}>\right)(\mathbf{u}),\ \mathcal{L}_{\mathrm{eik}} = \frac{1}{K} \sum_{i=1}^K (\Vert \nabla_{\mathbf{x}} d(\mathbf{x}_i) \Vert_2 - 1 )^2,
\end{aligned}
\end{equation}
where $\mathcal{L}_{\mathrm{color}}$ and $\mathcal{L}_{\mathrm{mask}}$ are the radiance rendering loss and the BCE loss of object masks in NeuS \citep{neus}. Splatted 2D Gaussians is supervised by $\boldsymbol{\psi}(\mathbf{u})$ and $\widetilde{\mathbf{\Sigma}}_{\mathbf{p}}(\mathbf{u})$ in Eq. \ref{eq:2dpol}. $\hat{\boldsymbol{\varphi}}(\mathbf{n}_{\mathbf{p}}(\mathbf{u})) \equiv \boldsymbol{\psi}(\mathbf{u}) + \frac{\pi}{2} \mod \pi$ and $\boldsymbol{\psi}$ is the azimuth angle of normals. The supervision of radiance and polarization priors are reweighted by the DoP $\boldsymbol{\rho}$. Especially, only Anisotropy (ratio of singular values) and eigenvectors are supervised to avoid scaling and numerical issues. If the local shape is like a plane, normals will change smoothly in all directions, and the Anisotropy approaches $1$. If there are some details like edges, normals tend to change abruptly and exhibit directionality, represented by eigenvectors. $\mathcal{L}_{\mathrm{eik}}$ is a regularization term of the gradient of SDF widely used \citep{eik}. $\alpha$, $\beta$, $\gamma$ and $\delta$ are hyper-parameters.
\section{Experiments}
To evaluate the effectiveness of our method, we tested GNeRP on objects from multiple scenes and compared them with existing state-of-the-art neural 3D reconstruction methods.

\paragraph{PolRef Dataset} The methods are evaluated on the PANDORA dataset \citep{pandora} and captured scenes by ourselves. The PANDORA includes $3$ reflective objects (Owl, Blackvase, and Gnome) with polarization priors. However, their ground truth shapes are unavailable for quantitative evaluation. Moreover, the diversity of materials, geometry, and illumination is not enough for overall comparisons. Only the geometry of the Gnome scene is complicated but less reflective. Only a mirror-like ball in the Blackvase reflects surroundings other than highlights. Other common datasets, including Shiny Blender \citep{ref-nerf}, lack polarization priors for our method.  To comprehensively evaluate the performance of 3D reconstruction methods, a new challenging multi-view dataset named PolRef was collected, consisting of objects with reflective and less-textured surfaces captured with various illumination. Radiance images and aligned polarization priors were captured in one shot using polarization cameras. To obtain precise and complete ground truth shapes, objects were produced using SLA 3D printers, with an accuracy tolerance of $\pm 0.1 mm$. Detailed descriptions are shown in the Appendix. The dataset will be released to facilitate further research on 3D reconstruction in more challenging scenes in the future.
\paragraph{Experimental Settings} GNeRP is built upon NeuS \citep{neus}. The geometry network and radiance network in Fig. \ref{fig:net} is the same as that of NeuS. Since the covariance loss $\mathcal{L}_{\mathrm{cov}}$in Eq. \ref{eq:loss} refines the details of the geometry, it will not be activated during the initial $50$K steps. The model is trained for 200k iterations and takes about $6$ hours on a server with $4$  NVIDIA RTX 3090 Ti GPUs for the reconstruction. After optimization, the meshes are extracted from learned SDF by Marching Cubes \citep{marchcube} with a resolution of $512^3$. The hyper-parameter settings are shown in the Appendix.
\subsection{Comparison with State-of-the-Art Methods}
We conducted the comparison of reconstruction accuracy between our methods and several state-of-the-art methods, including baseline methods for neural 3D reconstruction (Unisurf \citep{unisurf}, VolSDF \citep{volsdf}, and NeuS \citep{neus}), two view-consistency based methods (NeuralWarp \citep{neuwarp} and Geo-NeuS \citep{geoneus}), two new methods for reconstruction of reflective objects (NeRO \citep{nero} and Ref-NeuS \citep{refneus}), and a polarization-based method (PANDORA \citep{pandora}).

\begin{figure}[h]
\centering
\captionsetup[subfloat]{labelsep=none,format=plain,labelformat=empty}
\makebox[\textwidth][c]{\subfloat{  
		\includegraphics[width=1\textwidth]{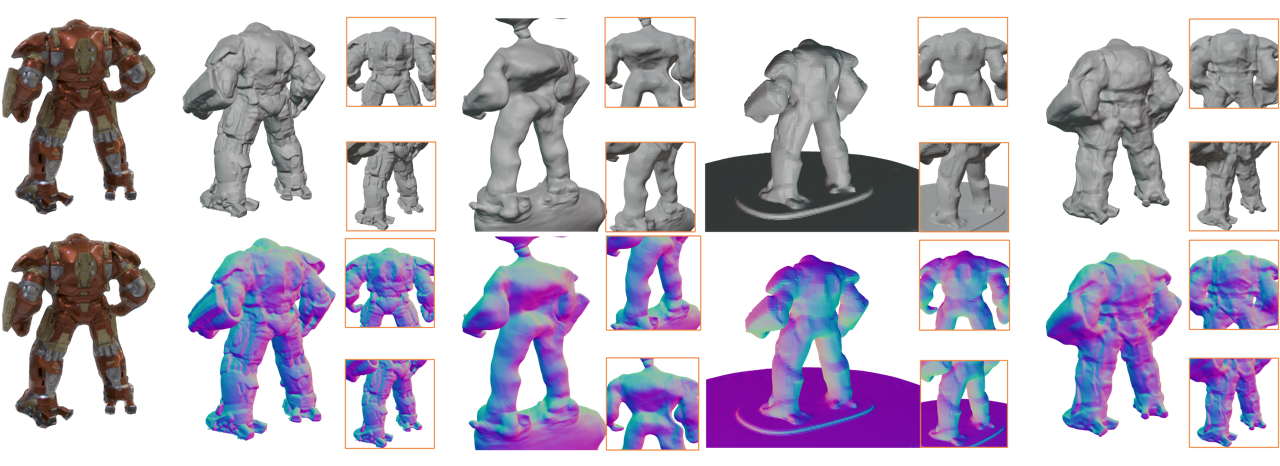}}}\\
\makebox[\textwidth][c]{\subfloat{  
		\includegraphics[width=1\textwidth]{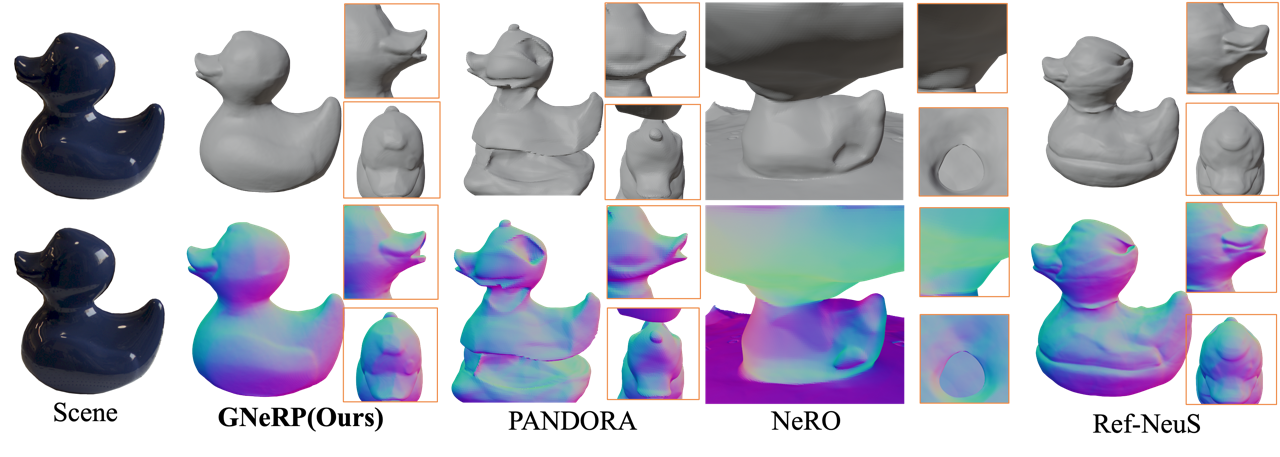}}}

\caption{Visual comparison of our method and state-of-the-art methods.}
\label{fig:vcomp}
\end{figure}

A qualitative comparison between our method and state-of-the-art methods specially designed for reflective objects is shown in Fig. \ref{fig:vcomp}, which demonstrates that our method significantly improves the geometry details and accuracy of normals. In the Ironman scene, NeRO reconstructed an over-smoothed geometry. Due to the spatial continuity of neural BRDF, it failed to reconstruct the high-frequency armor details with abrupt normal changes. The shape of Ref-NeuS is more accurate, but the sole scalar SDF is not able to predict the geometry details. The duck scene is more reflective with a combination of highlights and reflection of surroundings. Although Ref-NeuS detected the reflective regions, it was still misled by the environment radiance and reconstructing concave holes. The results of PANDORA are over-smoothed in Ironman and disturbed by noise in polarization priors in Duck. Additional comparisons of different scenes are shown in the Appendix.
\begin{table}[h]
\begin{center}
\begin{tabular}{lccccc}
\multicolumn{1}{c}{\bf Methods}  &\multicolumn{1}{c}{\bf Ironman} &\multicolumn{1}{c}{\bf Duck} &\multicolumn{1}{c}{\bf Cow} &\multicolumn{1}{c}{\bf Snorlax} &\multicolumn{1}{c}{\bf Mean} 
\\ \toprule \\
\multicolumn{1}{c}{Unisurf* \citep{unisurf}}  &\multicolumn{1}{c}{3.97}&\multicolumn{1}{c}{10.83}&\multicolumn{1}{c}{14.43}&\multicolumn{1}{c}{14.33}&\multicolumn{1}{c}{10.89} \\
\multicolumn{1}{c}{VolSDF \citep{volsdf}}    &\multicolumn{1}{c}{2.72}&\multicolumn{1}{c}{5.16}&\multicolumn{1}{c}{5.95}&\multicolumn{1}{c}{3.20}&\multicolumn{1}{c}{4.26}  \\
\multicolumn{1}{c}{NeuS  \citep{neus} }      &\multicolumn{1}{c}{\underline{2.28}}&\multicolumn{1}{c}{\underline{2.12}}&\multicolumn{1}{c}{3.82}&\multicolumn{1}{c}{\underline{2.11}}&\multicolumn{1}{c}{\underline{2.58}} \\
\multicolumn{1}{c}{Geo-NeuS \citep{geoneus}}&\multicolumn{1}{c}{4.77}&\multicolumn{1}{c}{10.12}&\multicolumn{1}{c}{17.48}&\multicolumn{1}{c}{10.39}&\multicolumn{1}{c}{10.82}\\
\multicolumn{1}{c}{NeuralWarp \citep{neuwarp}}&\multicolumn{1}{c}{12.44}&\multicolumn{1}{c}{19.78}&\multicolumn{1}{c}{5.41}&\multicolumn{1}{c}{20.57}&\multicolumn{1}{c}{14.55}\\
\multicolumn{1}{c}{NeRO* \citep{nero}}&\multicolumn{1}{c}{2.29}&\multicolumn{1}{c}{23.75}&\multicolumn{1}{c}{\underline{\underline{2.95}}}&\multicolumn{1}{c}{24.30}&\multicolumn{1}{c}{13.32}\\
\multicolumn{1}{c}{Ref-NeuS \citep{refneus}}&\multicolumn{1}{c}{\underline{\underline{1.88}}}&\multicolumn{1}{c}{\underline{\underline{1.93}}}&\multicolumn{1}{c}{\underline{3.66}}&\multicolumn{1}{c}{\underline{\underline{1.99}}}&\multicolumn{1}{c}{\underline{\underline{2.34}}}\\
\multicolumn{1}{c}{PANDORA $^{\dagger}$ \citep{pandora}}&\multicolumn{1}{c}{4.61}&\multicolumn{1}{c}{5.28}&\multicolumn{1}{c}{7.96}&\multicolumn{1}{c}{5.73}&\multicolumn{1}{c}{5.90} \\
\midrule
\multicolumn{1}{c}{GNeRP}&\multicolumn{1}{c}{\textbf{1.34}}&\multicolumn{1}{c}{\textbf{1.63}}&\multicolumn{1}{c}{\textbf{1.39}}&\multicolumn{1}{c}{\textbf{1.05}}&\multicolumn{1}{c}{\textbf{1.35}}
\\\bottomrule
\end{tabular}
\caption{Quantitative comparison with state-of-the-art methods. The lower is better. * indicates the method doesn't use object masks. $\dagger$ refers to the use of polarization priors. The best scores are \textbf{bold}, the second best scores are \underline{\underline{double underlined}}, and the third best scores are \underline{underlined}.}
\label{tab:benchmark}
\end{center}
\end{table}
We conduct quantitative comparisons on the four scenes with ground truth meshes in our dataset. The evaluation metric is Chamfer Distance according to NeuS \citep{neus} and Unisurf \citep{unisurf}. Scores are reported in Table \ref{tab:benchmark}, which shows our method reconstructs more precise meshes in all four scenes. NeRO \citep{nero} needs environment information to calculate occlusion loss, and Unisurf also learns occupancy from backgrounds. Training them with masks failed directly, so we report the scores without masks in Tab. \ref{tab:benchmark} denoted by *. Geo-NeuS needs sparse points from Structure-from-Motion (SfM) \cite{colmap} to calculate SDF loss and select the pairs based on SfM for the warping process. We did the sparse reconstruction in COLMAP and followed the pairs selection method in NeuralWarp. Full polarimetric acquisition (Stokes vector $[\mathbf{s}_0, \mathbf{s}_1, \mathbf{s}_2]$, see in the Appendix) is required by PANDORA. We processed the raw polarization capture to follow its data conventions. Sparse reconstruction of reflective scenes was noisy and incomplete, resulting in the worst accuracy by Geo-NeuS and NeuralWarp. 
Ref-NeuS demonstrated comparable scores on all scenes, but our method still outperformed. 
\subsection{Ablation Study}
\begin{table}[h]
\begin{center}
\begin{tabular}{lcccccc}
\multicolumn{1}{c}{Scene}  &\multicolumn{1}{c}{NeuS} &\multicolumn{1}{c}{w/ $\mathcal{L}_{\mathrm{mean}}$} &\multicolumn{1}{c}{w/ $\mathcal{L}_{\mathrm{cov}}$} &\multicolumn{1}{c}{w/ ReW. $\mathcal{L}_{\mathrm{mean}}$}&\multicolumn{1}{c}{w/ ReW. $\mathcal{L}_{\mathrm{cov}}$}&\multicolumn{1}{c}{Full}
\\ \hline \\
\multicolumn{1}{c}{Snorlax} &\multicolumn{1}{c}{2.11} &\multicolumn{1}{c}{2.03} &\multicolumn{1}{c}{3.01}&\multicolumn{1}{c}{1.81} &\multicolumn{1}{c}{2.07}&\multicolumn{1}{c}{\textbf{1.05}} \\
\multicolumn{1}{c}{Cow} &\multicolumn{1}{c}{3.82}&\multicolumn{1}{c}{2.72}  &\multicolumn{1}{c}{5.54}&\multicolumn{1}{c}{1.94} &\multicolumn{1}{c}{2.29} &\multicolumn{1}{c}{\textbf{1.39}} 
\\\hline
\end{tabular}
\caption{Ablation Study. $\mathcal{L}_{\mathrm{mean}}$, $\mathcal{L}_{\mathrm{cov}}$ are in Eq. \ref{eq:loss}.}
\label{tab:abla}
\end{center}
\end{table}

To validate the effectiveness of the proposed modules, we test the following three settings as shown in Tab. \ref{tab:abla}. W/$\mathcal{L}_{\mathrm{mean}}$ refers to the naive supervision of $\mathbf{\varphi}$ and azimuth angle of normals in SDF. Due to the noise, the results are worse. W/$\mathcal{L}_{\mathrm{mean}}$ refers to the polarization supervision with only covariance. The reconstruction is focused on details and results in the worst scores. W/ ReW. $\mathcal{L}_{\mathrm{mean}}$ indicates the reweighted losses $(1-\boldsymbol{\rho})\mathcal{L}_{\mathrm{color}} +\boldsymbol{\rho}\mathcal{L}_{\mathrm{mean}}$. Similarly, w/ReW. $\mathcal{L}_{\mathrm{cov}}$ represents $(1-\boldsymbol{\rho})\mathcal{L}_{\mathrm{color}} +\boldsymbol{\rho}\mathcal{L}_{\mathrm{cov}}$. The reweighting does improve the efficiency of polarization priors. Finally, the full setting shows the best scores. Additional visualization is shown in the Appendix.
\section{Conclusion}
We propose GNeRP to reconstruct the detailed geometry of reflective scenes. In GNeRP, we propose a new Gaussian-based representation of normals and introduce polarization priors to supervise it. We propose a DoP reweighing strategy to resolve noise issues in polarization priors. We collect a new, challenging multi-view dataset with non-Lambertian scenes to evaluate existing methods more comprehensively. Experimental results demonstrate the superiority of our method.

\bibliography{iclr2024_conference}
\bibliographystyle{iclr2024_conference}

\end{document}